\def\year{2015}
\documentclass[letterpaper]{article}
\usepackage{aaai}
\usepackage{times}
\usepackage{helvet}
\usepackage{courier}
\usepackage{graphicx}
\usepackage{amsmath}
\usepackage{amssymb}
\usepackage{subcaption}
\usepackage{paralist}
\usepackage{algorithm}
\usepackage{algpseudocode}
\usepackage{amsthm}
\usepackage{makecell}
\usepackage{enumitem}
\usepackage{xcolor}
\usepackage{url}

\newcommand{\change}[1]
		{#1}

\newcommand\note[1]{}

\nocopyright

\newtheorem{lemma}{Lemma}

\begin{document}
 \setcounter{secnumdepth}{2}

%
\title{Are Elephants Bigger than Butterflies?\\ Reasoning about Sizes of Objects}
\author{Hessam Bagherinezhad$^\dagger$ \and Hannaneh Hajishirzi$^\dagger$ \and Yejin Choi$^\dagger$ \and Ali Farhadi$^\dagger$$^\ddagger$\\
$^\dagger$University of Washington, $^\ddagger$Allen Institute for AI \\
\{hessam, hannaneh, yejin, ali\}@washington.edu
}
\maketitle

\begin{abstract}
Human vision greatly benefits from the information about sizes of objects. The role of  size in several visual reasoning tasks has been thoroughly explored in human perception and cognition. However, the impact of the information about sizes of objects is yet to be determined in AI. We postulate that this is mainly attributed to the lack of a comprehensive repository of size information. In this paper, we introduce a method to automatically infer object sizes, leveraging visual and textual information from web.  By maximizing the joint likelihood of textual and visual observations, our method learns reliable relative size estimates, with no explicit human supervision.  We introduce the relative size dataset and show that our method outperforms  competitive textual and visual baselines in reasoning about size comparisons.

\end{abstract}

\section{Introduction}
Human visual system has a strong prior knowledge about physical sizes of objects in the real world~\cite{ittelson1951size} and can immediately retrieve size information as it recognizes objects~\cite{konkle2012familiar}. Humans are often very sensitive to discrepancies in size estimates (size constancy~\cite{holway1941determinants}) and draw or imagine objects in canonical sizes, despite significant variations due to a change in viewpoint or distance~\cite{konkle2011canonical}.
Considering the importance of size information in human vision, it is counter-intuitive that most of the current AI systems 
are agnostic to  object sizes. We postulate that this is mainly due to the lack of a comprehensive resource that can provide information about  object sizes. 
 In this paper, we introduce a method to automatically provide such information by representing and inferring object sizes and their relations. To be comprehensive, our method does not rely on explicit human supervision and only uses web data.

\begin{figure}[t]
\includegraphics[width=0.44\textwidth]{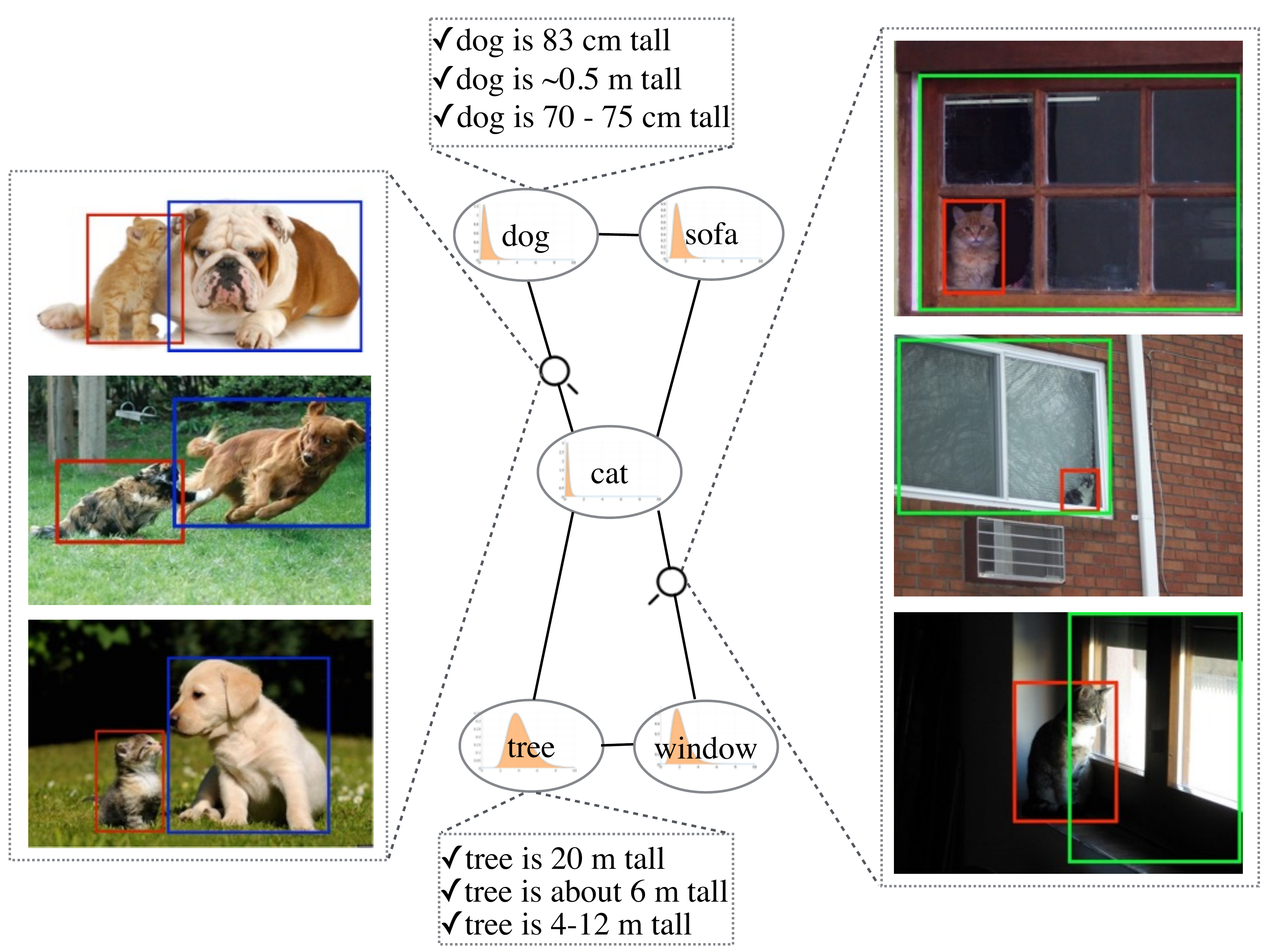}\vskip -.3cm
\caption{\small In this paper we study the problem of inferring sizes of objects using visual and textual data available on the web. With no explicit human supervision, our method achieves reliable ($83.5$\% accurate) relative size estimates. We use size graph, shown above, to represent both absolute size information (from textual web data) and relative ones (from visual web data). The size graph allows us to leverage the transitive nature of size information by maximizing the likelihood of both visual and textual observations.} \vskip -.2cm
\label{fig:teaser}
\end{figure}

Identifying numerical properties of  objects, such as size, has been recently studied in Natural Language Processing and shown to be helpful for question answering and information extraction~\cite{nlp-comparative,Chu-carroll03,nlp-hebrew}. The core idea of the state-of-the-art methods is to design search queries in the form of  manually defined templates either looking for absolute size of objects (e.g. ``the size of a car is * unit'') or specific relations (e.g. ``wheel of a car''). The results are promising,  but the quality and scale of such extraction has been somewhat limiting.  For example, these methods predict a relatively small size for a `car' because search queries discover more frequent relations about the size of a `toy car' rather than a regular `car'~\cite{nlp-tokyo}.
 This is in part because  most trivial commonsense knowledge is rarely stated explicitly in natural language text, e.g., it is unlikely to find a sentence that says a car is bigger than an orange. In addition, comparative statements in text, if found, rarely provide precisely how much one object is bigger than the other. In this paper, we argue that visual and textual observations are complementary, and a successful size estimation method will take advantage of both modalities.

In images, estimating the absolute sizes of  objects requires information about the camera parameters and accurate depth estimates which are not available at scale.  Visual data, however, can provide informative cues about relative sizes of objects. For example, consider the `cat' that is sitting by the `window' in Figure~\ref{fig:teaser}. The relative size of the `cat' and the `window' can be computed using their detection boxes, adjusted by their coarse depth.  A probability distribution over relative sizes of `cats' and `windows' can then be computed by observing several images in which `cats' and `windows' co-occur. However,  not all pairs of objects appear in large enough number of images. Collecting visual observations for some pairs like `sofa' and `tree' is not possible.  Furthermore, it is not scalable  to collect visual observations for all pairs of objects. 



 In this paper, we introduce a method to learn to estimate sizes of objects, with no explicit human supervision,  leveraging both textual and visual observations. 
 Our approach is to couple (noisy) textual and visual  estimates and use the transitive nature of size information to reason about objects that don't co-occur frequently. For example in Figure~\ref{fig:teaser},  our method can establish inferences about the relative size of `sofa' and `tree' through a set of intermediate relations between `sofa'-`cat' and `cat'-`tree'.  
 

We introduce {\it size graph} as our representation to model object sizes and their relations.  The nodes in the size graph correspond to the log-normal distribution of the sizes of objects and edges correspond to relative sizes of pairs of objects that co-occur frequently. The topology of the size graph provides guidance on how to collect enough textual and visual observations to deal with the noise and sparsity of the observations.
We formulate the problem of learning the size of the objects as optimizing for a set of parameters that maximize the likelihood of both textual and visual observations. To obtain large scale visual observations we use detectors trained without explicit annotations using webdata~\cite{levan} and single image depth estimators that are pretrained using few categories and have shown to be generalizable to unseen categories.  

Our experimental evaluations show strong results. On our dataset of about $500$ relative size comparisons, our method achieves $83.5$\% accuracy, compared to $63.4$\% of a competitive NLP baseline. Our results show that textual and visual data are complementary, and optimizing for both outperforms individual models. If available, our model can benefit from reliable information about the actual sizes of a limited number of object categories. \footnote{ The code, data, and results are available at \change{{\small\url{http://grail.cs.washington.edu/projects/size}}}.}


\vspace{-2mm}
\section{Related Work}


A few researchers~\cite{pragerCCWIM03,Chu-carroll03} use manually curated commonsense knowledge base such as OpenCyc~\cite{opencyc} for answering questions about numerical information. 
These knowledge resources (e.g., ConceptNet~\cite{conceptnet})  usually consist of taxonomic assertions or generic relations, but do not include size information. Manual annotations of such knowledge is not scalable.   Our efforts will result in extracting size information to populate such knowledge bases (esp. ConceptNet) with size information at scale.

Identifying numerical attributes about objects has been addressed in NLP recently. The common theme in the recent work~\cite{nlp-tokyo,nlp-hebrew,Iftene10,nlp-comparative,nlp-recent} is to use search query templates with other textual cues  (e.g., more than, at least, as many as, etc), collect numerical values, and model sizes as a normal distribution. 
However, the quality and scale of such extraction is somewhat limiting. 
Similar to previous work that show textual and visual information are complementary across different domains~\cite{seo2015solving,neil,spt}, we show that a successful size estimation method should also take advantage of both modalities. In particular, our experiments show that textual observations about the relative sizes of objects are very limited, and relative size comparisons are better collected through visual data. In addition, we show that log-normal distribution is a better model for representing sizes than normal distributions.

In computer vision, size information manually extracted from furniture catalogs, has shown to be effective in indoor scenes understanding and reconstruction~\cite{pero2012bayesian}. However, size information is not playing a major role in mainstream computer vision tasks yet. This might be due to the fact that there is no unified and comprehensive resource for objects sizes. The visual size of the objects depends on multiple factors including the distance to the objects and the viewpoint.  Single image depth estimation has been an active topic in computer vision~\cite{delage2006dynamic,hedau2009recovering,liu2010single,saxena2005learning,ladicky2014pulling}. 
In this paper, we use~\cite{Deepdepth} for single image depth estimation.

\vspace{-2mm}
\section{Overview of Our Method}
\noindent{\bf Problem Overview: }
In this paper, we address the problem of identifying sizes of physical objects using visual and textual information. Our goals are to (a) collect visual observation about the relative sizes of objects,  (b) collect  textual observations about the absolute sizes of objects,  and (c)  devise a method to make sense of vast amount of visual and textual observations and estimate object sizes. We evaluate our method by answering queries about the size comparisons: if the object \texttt{A} is bigger than the object \texttt{B} for every two objects \texttt{A} and \texttt{B} in our dataset. 

\begin{algorithm}
\small
\caption{The overview of our method.}
\begin{algorithmic}[1]
\State {\bf Representation}: Construct   Size Graph (Section~\ref{subsec:graph}). 
\State  \Comment{Collect Visual observations (Section~\ref{subsec:collection})\ \ \ \ \ \ \ \ \ \ \ \ \ \ \ \ \ \ \   }
\For{all edges $(v,u)$ in the  Size Graph 
}
\State Get images from Flickr in which $v$ and $u$ are tagged.
\State Run object detectors of $v$ and $u$ on all images.
\State Observe the depth adjusted ratio of bounding box areas.
\EndFor
\State  \Comment{Collect Textual observations (Section~\ref{subsec:collection})\ \ \ \ \ \ \ \ \ \ \ \ \ \ \ \ \ \ \   }
\For{all nodes $v$ in the  Size Graph}
\State Execute search engine patterns for each object.
\State Observe the sizes found for objects.
\EndFor
\State Model the size of each object with a log-normal.
\State {\bf Learning:} Find the optimal parameters maximizing the likelihood (Section~\ref{subsec:learning}).
\end{algorithmic}
\label{alg:overview}
\end{algorithm}\vspace{-4mm}

\noindent{\bf Overview of Our Method: } 
We devise a method (Algorithm~\ref{alg:overview}) that learns probability distributions over object sizes based on the observations gathered from both visual and textual web, with no explicit human supervision.    In order to deal with the noise and incompleteness of the data,  we introduce {\it size graph} that represents object sizes (nodes) and their relations (edges) in a connected, yet sparse graph representation (Section~\ref{sec:representation}). 

 We use textual web data to extract information about the absolute sizes of objects through search query templates. We use web images to extract information about the relative sizes of objects if they co-occur in an image. With scalablity in mind,   we incorporate webly-supervised object detectors \cite{levan} to detect the objects in the image and compute the depth adjusted ratio of the areas of the detected bounding boxes for objects (Section~\ref{subsec:collection}). 
 
We formulate the problem of estimating the size as  maximizing the likelihood of textual and visual observations to learn  distributions over object sizes  (Section~\ref{subsec:learning}). Finally, we incorporate an inference algorithm to answer queries in the form of  ``Which object is bigger?" (Section~\ref{subsec:inference}). 

\vspace{-2mm}
\section{Representation: Size Graph}
\label{sec:representation}
It is not scalable to collect visual observations for all pairs of objects. In addition, for some pairs like `aeroplane' and `apple', it is noisy (if at all possible) to directly collect visual observations.  
We introduce {\it size graph} as a compact, well-connected,  sparse graph representation (Section~\ref{subsec:graph}) whose nodes are distributions over the actual sizes of the objects (Section~\ref{subsec:lognormal}). The properties of the size graph allows us to collect enough visual and textual data suitable for modeling the size distributions.  
\vspace{-1mm}
\subsection{Graph Construction}\label{subsec:graph}

\noindent{\bf Size Graph Properties:}  Given a list of objects $V = \{O_1, O_2, \cdots, O_n\}$, we want to construct a graph $G = (V,E)$ such that there is one node for every object and there exists an edge $(O_i$, $O_j) \in E$ only if $O_i$ and $O_j$ {\it co-occur} frequently in images. In particular, the size graph should have the following properties:\begin{inparaenum}[(a)]\item Connectivity, which allows us to take advantage of the transitivity of size and propagate any size information throughout the graph. 
In addition, we require that there are at least $k$ disjoint paths between every two nodes in the graph in order to reduce the effect of noisy edges in the graph. \item Sparsity, which allows us to collect enough visual data since it is not feasible (both computationally and statistically) to connect every two nodes in the graph. Adding an edge between two unrelated objects like `apple' and `bicycle' not only increases the computational cost, but also increases the noise of the observations. \end{inparaenum} 



\vskip .1cm
\noindent{\bf Modeling Co-occurrence:}  We approximate the likelihood of co-occurrence of two objects in images using the tag lists of images in Flickr 100M dataset. Every image in Flickr is accompanied with a list of tags including names of objects. We use the co-occurrence of two objects in tag lists of Flickr images as a proxy for how much those objects are likely to co-occur in images. We observed that not all co-occurrences are equally important and shorter tag lists are more descriptive (compared to longer lists). We first define the descriptiveness of a tag list as the inverse of the length of the list. Then, we compute co-occurrence of objects $O_i$ and $O_j$ by summing over the descriptiveness of the tag lists in which both objects $O_i$ and $O_j$ co-occur.

We define the cost $\epsilon_{ij}$ of an edge $e_{i,j} = (O_i, O_j)$ in the complete graph as the inverse of the co-occurrence of $O_i$ and $O_j$. Therefore, if two objects co-occur frequently in a short list of tags, the cost of an edge is small.  Let $L_l$ be the tag list of the $l_{th}$ image in Flickr 100M dataset, the following equation formulates the cost of an edge $(O_i, O_j)$:
\vspace{-1.5mm}
\begin{equation}
\label{eq:tfidf}
\epsilon_{ij} =
\begin{cases}
\frac{1}{\sum_{l: \{O_i, O_j\} \subseteq L_l} \frac{1}{|L_l|}},& \text{if } \exists k: \{O_i, O_j\} \subseteq L_l\\
\infty,                                                                        & \text{otherwise}
\end{cases}
\end{equation}\vspace{-3mm}

\noindent{\bf Constructing Size Graph: } Let $D$ be the weighted complete graph of objects, with edge costs define by equation~\ref{eq:tfidf}. According to the properties of the size graph,  our goal is to find a minimum cost subgraph of $D$ in which there are multiple disjoint paths between every two nodes. Such subgraph would be less susceptible to the noise of visual observations across edges. As a corollary to Menger's theorem~\cite{menger}, there are at least $k$ disjoint paths between every two nodes of an arbitrary graph $G$ if and only if $G$ is $k$-edge-connected (if we remove any $k-1$ edges, the graph is still connected). Therefore, our goal here is to  find the minimum $k$-edge-connected subgraph.
%
The problem of finding the minimum $k$-edge-connected subgraph, however, is shown to be NP-hard for $k > 1$~\cite{NPbook}. 

Here, we introduce our algorithm to find a $k$-edge-connected subgraph whose cost is an approximation of the optimal cost. Our approximation algorithm is to iteratively find a minimum spanning tree (MST) $T_1 \subseteq D$, and remove its edges from $D$, and then continue with finding another MST of the remaining graph. Repeating this iteration for $k$ times results in $k$ disjoint spanning trees  $T_1, T_2, \cdots, T_k$. The final subgraph $G = T_1 \cup \cdots \cup T_k$ is then derived by combining all these spanning trees together. The subgraph $G$ is $k$-edge-connected, and its cost is an approximation of the optimal cost. 

\begin{lemma}
\label{lem:k-edge}
Every graph $H = T_1 \cup \cdots \cup T_k$ which is a union of $k$ disjoint spanning trees is $k$-edge-connected.
\end{lemma}\vspace{-.3cm}
\begin{proof}
In order to make $H$ disconnected, at least one edge should be removed from each spanning tree. Since spanning trees are disjoint, at least $k$ edge removals are required to disconnect the graph $H$.
\end{proof}
\vspace{-.2cm}
\begin{lemma}
\label{lem:approx-factor}
Given a graph $G=(V,E)$, and the subgraph $H = T_1 \cup \cdots \cup T_k$ where $T_i$ is the $i_{th}$ MST of $G$. The total cost of $H$ is at most $\frac{2M}{m}$ times the cost of the optimal $k$-edge-connected subgraph, where $m$ and $M$ are the minimum and the maximum of edge costs, respectively.
\end{lemma}\vspace{-.1cm}
\begin{proof}
Let $OPT$ denote the optimal $k$-edge-connected subgraph. The minimum degree of $OPT$ should be at least $k$. Hence, $OPT$ must have at least $\frac{nk}{2}$ edges, each of which with the cost of at least $m$. Therefore $\frac{nkm}{2} \leq cost(OPT)$. On the other hand, the subgraph $H$ has exactly $k(n-1)$ edges, each of which with the cost of at most $M$. Hence, 
\vspace{-2mm}
\[   
\begin{split}
cost(H) \leq kM(n-1) < kMn = \frac{2M}{m} \times \frac{nkm}{2} \\
\leq \frac{2M}{m} cost(OPT) 
\end{split}
\] \vskip -8mm
\end{proof}

\vspace{-.1cm}
\subsection{Log-normal Sizes}
\label{subsec:lognormal}

There are many instances of the same object in the world, which vary in size. 
In this paper, we argue that the sizes of object instances are taken from a log-normal distribution specific to the object type i.e.,  the logarithm of sizes are taken from a normal distribution. This is different from what has been used in the previous work in NLP~\cite{nlp-hebrew} where the sizes of objects are  from a normal distribution. 

Let's assume the actual size of an apple comes from a normal distribution with $\mu = 5$ and $\sigma = 1$. The first problem  is a non-zero pdf for  $x\leq 0$, but physical objects cannot have negative sizes (probability mass leakage). The second problem is that the probability of finding an apple with a size less than $0.1$ ($\frac{1}{50}$ of an average apple) is greater than finding an apple with a size greater than $10$ (twice as big as an average apple), which is intuitively incorrect.
Using log-normal sizes would resolve both issues. Assume size of an apple comes from a log-normal distribution with parameters $\mu = \ln 5$ and $\sigma = 1$. With this assumption, the probability of finding an apple of negative size is zero. Also, the probability of finding an apple twice as big as an average apple is equal to seeing an apple whose size is half of an average apple. 

It is very interesting to see that the log-normal representation is aligned well with recent work in psychology that shows the visual size of the objects correlates with the log of their assumed size~\cite{konkle2011canonical}. In addition, our experimental results demonstrate that the log-normal representation improves the previous work.


\section{Learning Object Sizes}

\begin{figure*}[t]
\includegraphics[width=\textwidth]{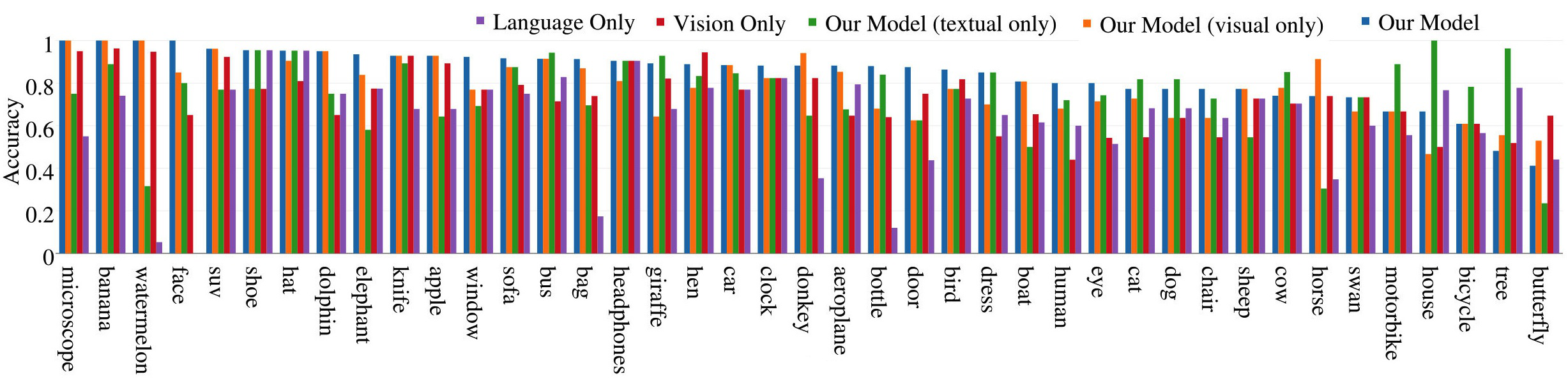}\vskip -3mm
\caption{\small The accuracy of models for objects in our dataset. Objects are sorted by the accuracy of our model.}\vskip -6mm
\label{fig:perobject}
\end{figure*}


\subsection{Collecting Observations}
\label{subsec:collection}


{\it Visual Observations: }We collect visual data to observe instances of relative sizes of objects. For each edge $e=(O_i, O_j)$ in the size graph, we download multiple images from Flickr that are tagged with both $O_i$ and $O_j$ and run the corresponding object detectors. These detectors  are trained by a webly-supervised algorithm~\cite{levan} to maintain scalability. Let $box_1$ and $box_2$ be the top predicted bounding boxes for the first and the second objects respectively. If the score of both predictions are above the default threshold of each detector, we record $r = \frac{area(box_1)}{area(box_2)} \times \frac{depth(box_1)^2}{depth(box_2)^2}$, as an observation for the relative size $\frac{size(O_i)}{size(O_j)}$. \change{Here,} $depth(box_i)$ is the average depth of $box_i$ computed from the depth estimation of \cite{Deepdepth}, \change{used according to Thales' theorem to normalize the object distances}. Note that our method does not use any bounding box information neither for detector training nor for depth estimation. We have used LEVAN~\cite{levan} detectors which are trained on google images with no human supervision. Depth estimator is pre-trained on Kinect data and has shown to generalize well for web images.

{\it Textual Observations:}  We collect textual data to observe instances of absolute sizes of objects.  In particular, we collect numerical values for the size of each object by executing search queries with the patterns of ``[object] * x * [unit]", ``[object] is * [unit] tall", and ``[object] width is * [unit]". These patterns are taken from previous works in the NLP community \cite{nlp-hebrew,nlp-tokyo}. Each search result might contain multiple numerical results. We compute the geometric mean of the multiple numerical values within each search result.  
 After scaling numerical results with respect to the unit used in each pattern we record them as observations for $size(O_i)$.
\subsection{Learning}
\label{subsec:learning}
As discussed in section \ref{subsec:lognormal}, we assume that $\log$ of object sizes comes from a normal distribution i.e.,  $g_i=\log size(O_i) \sim N(\mu_i, \sigma_i^2)$. The goal of the learning step is to find parameters $\mu_i$ and $\sigma_i$ for every object $O_i$ that maximizes the likelihood of the observations.	

Let $x_{ij}^{(r)}$ denote the $r_{th}$ binary visual observation for the relative size $\frac{size(O_i)}{size(O_j)}$,
 and let $x_{i}^{(r)}$ denote the $r_{th}$ unary textual observation for $size(O_i)$. We define  variables $y_{ij}^{(r)}= \log x_{ij}^{(r)}$ and $y_{i}^{(r)} = \log x_{i}^{(r)}$ as the logarithms of the observations  $x_{ij}^{(r)}$ and  $x_{i}^{(r)}$, respectively. This implies $y_i \sim g_i$  and $ y_{ij} \sim g_i-g_j$.  Assuming that the observations are independent, the log-likelihood of all observations is as follows: 
\vspace{-2mm}
\small
\begin{align}
\hspace{-2mm}\sum_{(i,j) \in E} \sum_{r=1}^{n_{ij}} \log f(g_i - g_j = y_{ij}^{(r)} | g_i \sim N(\mu_i, \sigma_i^2), g_j \sim N(\mu_j, \sigma_j^2))\notag&\\
&\hspace{-7.2cm}+ \sum_{i \in V} \sum_{r=1}^{n_i} \log f(g_i = y_{i}^{(r)} | g_i \sim N(\mu_i, \sigma_i^2)) \label{eq:loglike}
\end{align}
\normalsize
\noindent where $n_i$ is the number of textual observations for the $i$'th node, $n_{ij}$ is the total number of visual observations for the edge $(O_i, O_j)$, and $E$ is the set of edges in size graph. The first and the second summation terms of equation \ref{eq:loglike} refer to the log-likelihood of the visual and textual observations, respectively. 

We solve the above optimization by coordinate ascent. At each step we update parameters $\mu_i$ and $\sigma_i$ from the values of other parameters, assuming all the other parameters are fixed. For $\mu_i$ there is a closed form update rule; however, there is no closed form update for $\sigma_i$. To update $\sigma_i$, we do gradient ascent with the learning rate $\eta$.  The update rule for $\mu_i$ and $\sigma_i$, assuming all the other parameters are fixed are:
\vspace{-1.4mm}
\begin{align}
&\hspace{-7.5cm}\mu_i = \frac{\sum_{j: (i,j) \in E} \sum_{r=1}^{n_{ij}} \frac{y_{ij}^{(r)} + \mu_j}{\sigma_i^2 + \sigma_j^2} + \sum_{r=1}^{n_{i}} \frac{y_{i}^{(r)}}{\sigma_i^2}}{\sum_{j: (i,j) \in E} \frac{n_{ij}}{\sigma_i^2 + \sigma_j^2} + \frac{n_i}{\sigma_i^2}} \label{eq:updatemu}\\
\sigma_i^{(t+1)} = \sigma_i^{(t)} + \eta \bigg(\sum_{j:(i,j) \in E} \Big(\sum_{r=1}^{n_{ij}} \frac{\sigma_i^{(t)}(y_{ij}^{(r)} + \sigma_j - \sigma_i^{(t)^2})}{(\sigma_i^{(t)^2} + \sigma_j^2)} \notag &\\
&\hspace{-7cm}-\frac{n_{ij}\sigma_i^{(t)}}{\sigma_i^{(t)^2} + \sigma_j^2}\Big) + \sum_{r=1}^{n_i} \frac{(y_i^{(r)} - \mu_i)^2}{\sigma_i^{(t)^3}} - \frac{n_i}{\sigma_i^{(t)}}\bigg) \label{eq:updatesigma} 
\end{align}
The log likelihood (equation~\ref{eq:loglike}) is not convex. As a result, the coordinate ascent converges to a local optima depending on the initialization of the parameters. The non-convexity is due to the first summation; the second summation is convex. In practice, we initialize $\mu_i$ and $\sigma_i$ with the mean and the standard deviation of $Y_i = \{y_i^{(r)} | 1 \leq r \leq n_i\}$, which maximizes the second summation.


\subsection{Inference}
\label{subsec:inference}
After learning the parameters $\mu_i$ and $\sigma_i$ for all objects in our test set, we are able to infer if object $O_i$ is bigger than $O_j$  from the probability distributions of object sizes.
 Any linear combination of normal distributions is also a normal distribution; hence: 

\vspace{-0.5cm}
\small
\begin{equation*}
\vspace{-0.1cm}
\hspace{-2mm}
\begin{split}
\label{eq:inference}
P(size(O_i) > size(O_j)) =  P(\log size(O_i) - \log size(O_j) > 0)\\
= P(g_{ij} > 0 | g_{ij} \sim N(\mu_i - \mu_j, \sigma_i^2 + \sigma_j^2)) = 1 - \Phi(\frac{\mu_j - \mu_i}{\sqrt{\sigma_i^2 + \sigma_j^2}})
\end{split}
\end{equation*}
\normalsize
$\Phi(x)$ is the cumulative distribution function of the standard normal distribution and can be approximated numerically~\cite{cdf2,cdf4}. 


\section{Experiments}
We use Flickr 100M dataset~\cite{thomee2015yfcc100m} as the source of tag lists needed to construct the size graph (Section~\ref{subsec:graph}). We model size graph as a $2$-edge-connected subgraph since it is still sparse, the total cost of edges is small, and it does not get disconnected with the removal of an edge. 
 For each edge $(O_i, O_j)$ in the size graph, we retrieve a maximum of $100$ images from Flickr. We collect visual observations from the retrieved images and prune the outliers. To collect textual observations for the nodes, we execute our set of patterns on Google Custom Search Engine (Section~\ref{subsec:collection}). The code, data, and results can be found in the project website at \change{{\small\url{http://grail.cs.washington.edu/projects/size}}}
\subsection{Dataset}
It is hard, if possible, to evaluate our model with object categories absolute sizes, since there is no single absolute size for a category (i.e. the size of car varies from smallest mini cars to biggest SUVs). Therefore, we compiled a dataset of size comparisons among different physical objects. The dataset includes annotations for a set of object pairs $(O_i, O_j)$ for which people agree that $size(O_i) > size(O_j)$. The list of objects are selected from the $4869$ detectors in \textsc{LEVAN}~\cite{levan} that correspond to $41$ physical objects. To annotate the size comparisons, we deployed a webpage and asked annotators to answer queries of the form ``Which one is bigger, $O_i$ or $O_j$?" and possible answers include three choices of $O_i$, $O_j$, or `not obvious'. Annotators selected `not obvious' for non-trivial comparisons such as ``Which one is bigger, {\it bird} or {\it microscope}?".

We generated comparison surveys and asked each annotator $40$ unique comparison questions.  The annotators have shown to be consistent with each other on most of the questions (about 90\% agreement).  We only kept the pairs of objects that annotators have agreed and pruned out the comparisons with `not obvious' answers.  In total, there are  11 batches of comparison surveys and about 350 unique comparisons. To complete the list of annotated comparisons, we created a graph of all the available physical objects and added a directed edge from $O_i$ to $O_j$ if and only if people has annotated $O_i$ to be bigger than $O_j$. We verified that the generated graph is acyclic. We finally augmented the test set by adding all pairs of objects $(O_i, O_j)$ where there's a path from $O_i$ to $O_j$ in the graph.

Our final dataset includes a total of $486$ object pairs between $41$ physical objects. On average, each object  appears in  about $24$ comparison pairs where `{\it window}' with $13$ pairs has the least, and `{\it eye}' with $35$ pairs has the most number of pairs in the dataset.

\begin{figure*}
        \centering
        \begin{subfigure}[htbp]{0.35\textwidth} \vskip -4mm
        \small
        \centering
		\begin{tabular}{|l|c|}
			\hline
			Model & Accuracy \\
			\Xhline{4\arrayrulewidth}
			Chance & $0.5$\\
			\Xhline{3\arrayrulewidth}
			Language only & $0.634$\\
			\hline
			Vision only & $0.724$\\
			\Xhline{3\arrayrulewidth}
			Our model~(textual only) & $0.753$\\
			\hline
			Our model~(visual only) & $0.784$\\
			\hline
			Our model & $\mathbf{0.835}$\\
			\hline
		\end{tabular}
		\caption{\small The accuracy of our model against  baselines and ablations on estimating relative size comparisons. Our model outperforms competitive language-based and vision-based baselines by large margins. Our model benefits from both visual and textual information and outperforms language-only and vision-only ablations.}
		\label{tab:results}
	\end{subfigure}\hskip 5mm
        \begin{subfigure}[htbp]{0.25\textwidth} \vskip -2mm
                \includegraphics[width=\textwidth]{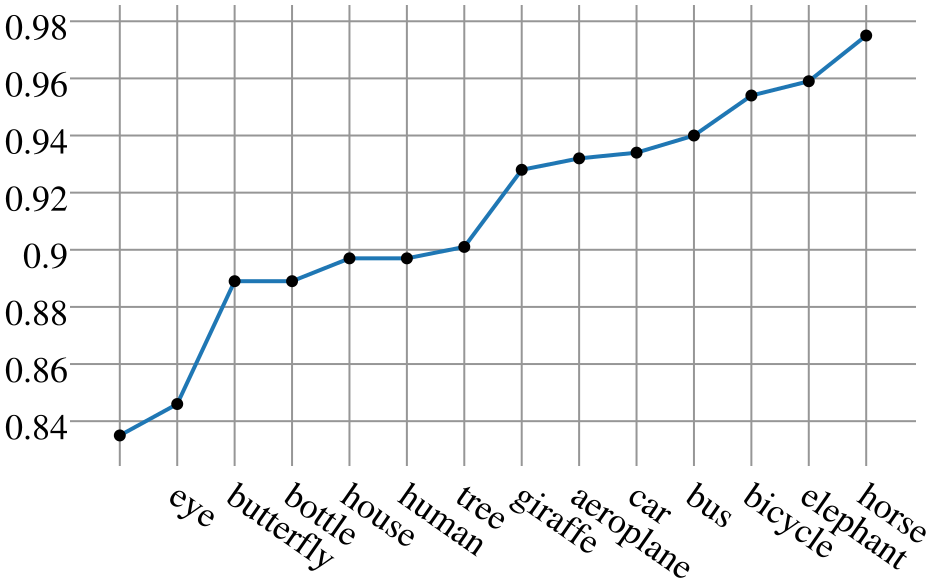}
			\caption{\small Our model can propagate information about true size of objects, if available. This figure shows an example case, where adding true estimates of the size information for about $10$ objects results in near perfect size estimates.}
			\label{fig:injection}
        \end{subfigure} \hskip 5mm
        \begin{subfigure}[htbp]{0.25\textwidth}
			\includegraphics[width=\textwidth]{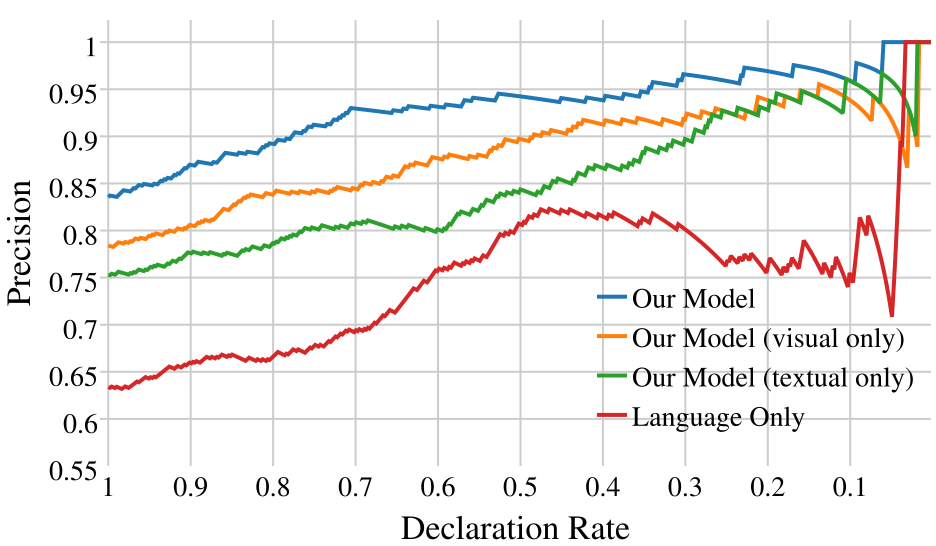}
			\caption{\small Precision vs. declaration rate in estimating the relative size information in our dataset. The curves are traced out by thresholding on $|P(A>B) - 0.5|$. Our model outperforms baselines in all declaration rates. }
			\label{fig:drate}
	   \end{subfigure}\vskip -3mm
	  \caption{}\vskip -7mm
\end{figure*}


\subsection{Comparisons}
\label{sec:baselines}
\noindent \textbf{Language-only baseline:} We re-implement~\cite{nlp-hebrew,nlp-tokyo} by forming and executing search engine queries with the size patterns mentioned in section~\ref{subsec:collection}. For every query, we record a size value after scaling the numerical results with respect to their units. The size of each object is then modeled with a normal distribution over observations.\footnote{ Our experiments have shown that textual observations about the relative sizes of physical objects are very limited. It is unlikely to find a sentence that says a car is bigger than an orange. In addition, comparative statements in text, if found, rarely provide precisely how much one object is bigger than the other.} 

\vskip .1cm
\noindent \textbf{Our model (textual only):} This is a variant of our model that only uses textual observations. This model maximizes the second production term of log likelihood (equation~\ref{eq:loglike}).

\vskip .1cm
\noindent \textbf{Vision-only baseline:} This is built on using the relative size comparisons directly taken from the visual data.  For each edge in the complete graph, we collect visual observations and set their relative size as the geometric mean of all the observations. To compute the relative size between any object pair, we multiply all the relative sizes of object pairs  in the shortest path between them.  

\vskip .1cm
\noindent \textbf{Our model (visual only):} This is a variant of our model that only uses visual observations. This model maximizes the first production term of log likelihood (equation~\ref{eq:loglike}). The difference between this model and vision-only baseline is on the representation (using size graph instead of complete graph) and also maximizing the likelihood, which involves observations altogether to estimate the objects' size distributions, instead of relying only on the shortest path edges.


\subsection{Results}

\noindent {\bf Overall Accuracy in Size Comparisons:} We report the accuracy of our model in inferring  size comparisons in  our dataset in Figure~\ref{tab:results}.  For inference, we compute $P(size(A) > size(B))$ (Section~\ref{subsec:inference}) and infer $A$ is bigger than $B$ if and only if $P(size(A) > size(B)) > 0.5$.  The accuracy is the number of correctly inferred pairs over all the pairs in the dataset. 

Our model achieves significant improvement over all the other models. The results confirm that visual and textual information are complementary and our model can take advantage of both modalities. In addition, our model~(textual only) achieves significantly higher performance compared to the language-only baseline.  This supports the superiority of our representation that sizes are represented with log-normal distributions. Finally, our model~(visual only) achieves significantly higher accuracy compared to the vision-only baseline. This confirms that maximizing the likelihood  removes the noise that exists in individual visual observations. 

\vskip .1cm
\noindent {\bf Per-object Accuracy: } Figure \ref{fig:perobject} shows that 
our model achieves higher accuracy than the baselines for most objects. For  objects like {\it giraffe}, {\it motorbike}, and {\it house} the textual data are less noisy and contribute more to the accuracy of our model, while for others like {\it watermelon}, {\it apple}, and {\it donkey} the visual data is more informative. 


\vskip .1cm
\noindent{\bf Precision vs. Declaration Rate:} 
All models (except the vision-only model) infer $A$ is bigger than $B$ if and only if $P(size(A) > size(B)) > 0.5$. We define the confidence of an estimation as the difference between the probability $P(size(A) > size(B))$ and $0.5$. 
Figure~\ref{fig:drate} shows the precision of the models vs. declaration rate~\cite{declarationRate}. Declaration rate is the proportion of the test queries on which the model outputs a decision. To calculate precision at a specific declaration rate $dr$, we first sort the queries in ascending order of each model's confidence, and then report precision over top $dr$ proportion of the test queries and discard the rest. Our results show that our model consistently outperforms other models at all declaration rates. It is worth mentioning that the precision of the language-only model drops at high confidence region ($dr > 0.5$), suggesting that the probabilistic model of this baseline is inaccurate.

\begin{figure}[t]
\centering
\includegraphics[width=0.4\textwidth]{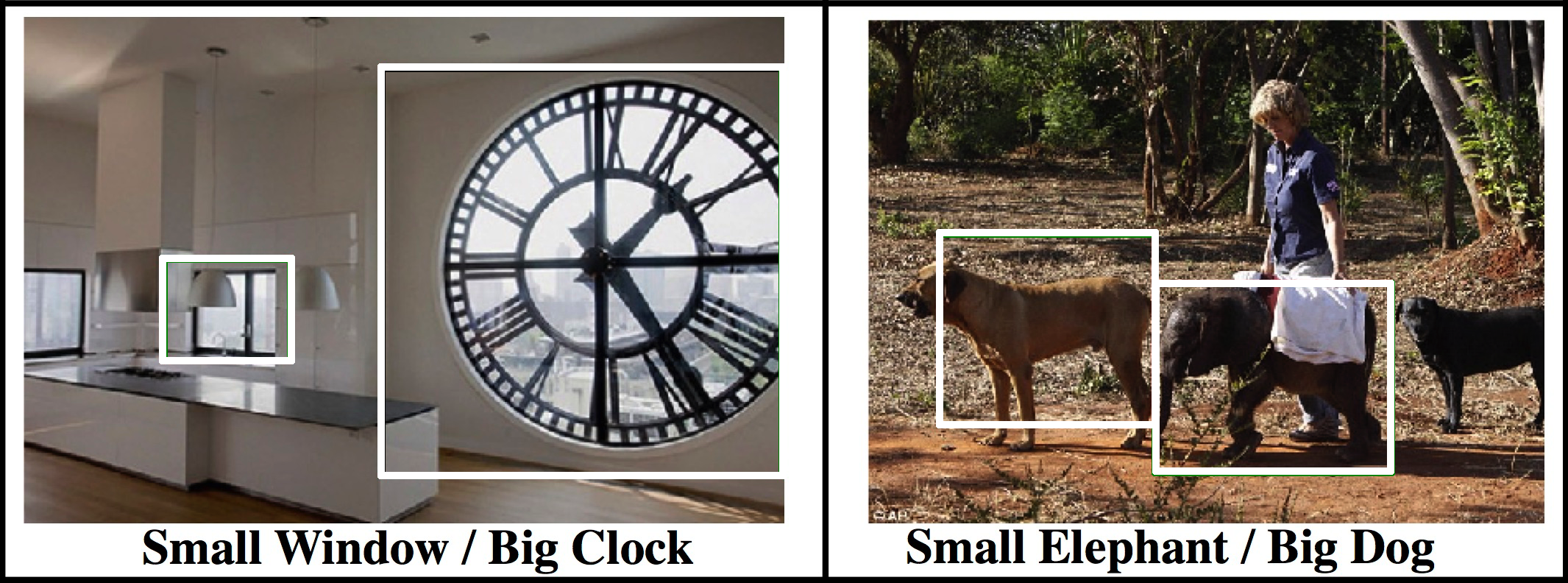}\vskip -.3cm
\caption{\small Relative size estimates can lead to inferences about atypical instances.}  
\label{fig:abnormal}\vspace{-5mm}
\end{figure}


\vskip .1cm
\noindent{\bf Sparse Supervision from True Sizes:} For a small number of objects, one might posses reliable size information. Our model can incorporate these information by fixing the size estimates for those objects and optimize the log-likelihood (equation~\ref{eq:loglike}) with respect to other objects' parameters. Our model is able to propagate information about the true object sizes to the uncertain nodes. 
Figure~\ref{fig:injection} shows the increase in accuracy when the true values of few objects are provided. 


\vskip .1cm
\noindent {\bf Qualitative Results:}
Size information is an important attribute for referring expressions and commonsense question answering~\cite{mitchell2011use,dialogQA} and can lead to inferences about  size abnormalities in images. For example, Figure \ref{fig:abnormal} shows examples of objects with unexpected relative size estimates. Rich statements, such as big clock/small window in Figure~\ref{fig:abnormal} can be used in image captioning or even pruning false positives in object detection. 

\change{The project website includes the size graph constructed using our method. The topology of the size graph reveals interesting properties about transitivity of the size information. For example, the size of chairs would be mainly affected by the estimates of the size of cats or the best way to  estimate the size of a sofa is through dogs and cats. Moreover, our method is able to estimate statistical size comparisons between objects which are not easy to compare by humans. For example, our method predicts that P(window$>$motorbike)=0.3, P(tree$>$SUV)=0.34, or P(shoe$>$face)=0.49. }
\vskip -1cm
\section{Conclusion}
In this paper, we introduced a fully automated method to infer information about sizes of objects using both visual and textual information available on the web. 
We evaluated our method on estimates of relative sizes of objects and show significant gain over competitive textual and visual baselines. We introduced size graph and showed its benefits in leveraging transitive nature of the size problem. Future work involves application of inferred size information in object detection in images and diagrams~\cite{seo2014diagram},  single image depth estimation, and building commonsense knowledge bases. This paper is a step toward the important problem of inferring the size information and can confidently declare that, yes, \textit{elephants are bigger than butterflies!}


\vskip .1cm
\noindent {\bf Acknowledgments:} This work was in part supported by ONR N00014-13-1-0720, NSF IIS-1218683, NSF IIS-1338054, and Allen Distinguished Investigator Award.

{
\small
\bibliographystyle{libs/aaai}

}
\end{document}